\documentclass[11pt]{article}

\usepackage[authoryear]{natbib}

\usepackage{graphicx}
\usepackage{amsmath,amssymb}
\usepackage{booktabs}
\usepackage{seqsplit}
\usepackage{subcaption}
\usepackage{siunitx}
\usepackage{xcolor}
\usepackage{hyperref}
\usepackage{xurl}
\usepackage{breakcites}

\title{Automatic Reflection Level Classification in Hungarian Student Essays}

\author{
Zsolt Csibi\textsuperscript{1}\thanks{Corresponding author: vrnf2j@inf.elte.hu}
\and
Kristian Fenech\textsuperscript{1}
\and
Mónika Sándor\textsuperscript{2}
\and
Mónika Serfőző\textsuperscript{2}
\and
Kinga Gyöngy\textsuperscript{2}
}

\date{
\textsuperscript{1}Department of Artificial Intelligence, Faculty of Informatics,\\
Eötvös Loránd University, Budapest, Hungary\\
\textsuperscript{2}Department of Education, Faculty of Primary and Pre-school Education,\\
Eötvös Loránd University, Budapest, Hungary\\[1ex]
\texttt{\{vrnf2j, fenech\}@inf.elte.hu}\\
\texttt{\{sandor.monika, serfozo.monika, gyongy.kinga\}@tok.elte.hu}
}

\newcommand{\cmark}{\textcolor{green!60!black}{\checkmark}}
\newcommand{\xmark}{\textcolor{red}{\texttimes}}

\begin{document}

% Short title for header
\maketitle
% Main content

\begin{abstract}
Reflective thinking is a key competency in education, but assessing reflective writing remains a 
time-consuming and subjective task for education experts. While automated reflective analysis has been explored in several 
languages, Hungarian language was not researched extensively. In this paper, we present the first comprehensive study on 
automatic reflection level classification in Hungarian student essays. We used a large, expert-annotated Hungarian dataset consisting of 1,954 reflective essays 
collected over multiple academic years and labeled on a four-level reflection scale. We investigate two 
approaches: (1) classical machine learning models using TF-IDF and semantic embedding features, 
and (2) Hungarian-specific transformer models fine-tuned for document-level reflection classification. 
To address the strong class imbalance in the dataset, we systematically examine class weighting, oversampling, 
data augmentation, and alternative loss functions. An extensive ablation 
study is conducted to analyze the contribution of each modeling and balancing strategy.
Our results show that shallow machine learning models with appropriate feature engineering achieve strong overall 
performance, reaching up to \textbf{71\%} overall score averaged over accuracy, F1-score, and ROC AUC metrics, while transformer-based models achieve slightly lower overall score (\textbf{68\%}) averaged over the same metrics,
but demonstrate better generalization on minority reflection classes. These findings highlight the continued relevance 
of classical methods for low-resource settings and the robustness of transformer models for imbalanced classification. 
The proposed dataset and experimental insights provide a solid foundation for future research on automated reflective 
analysis in Hungarian and other morphologically rich languages.
\end{abstract}


\maketitle

% Main content - import chapters
\section{Introduction}\label{sec:introduction}
Critically analyzing situations and trying to understand them better is a key skill in human cognition.
Reflective thinking is widely recognized as a core educational competency 
in international frameworks and policy documents \sloppy \citep{tuningacademy, oecd, europeanconuncil}. 
\sloppy
In higher education and professional training, reflective writing is commonly 
used to encourage deeper learning and professional development 
\citep{lee2005understanding, korthagen2022power, lim2023systematic, newton2000uncovering, occhiuto2024reflecting}. 
However, evaluating reflective texts is typically a manual process and therefore difficult 
to scale in educational practice \citep{dyment2011assessing, rogers2019validation, ullmann2019automated}.

To address this limitation, recent research has explored automatic reflective text 
analysis using both classical machine learning and transformer-based natural language 
processing approaches \citep{ullmann2019automated, alrashidi2022evaluating, alrashidi2024machine, wulff2023utilizing, zhang2024classification, nehyba2023applications, chong2020analysis}. 
These studies suggest that automated reflection classification is feasible across 
multiple settings and annotation schemes. At the same time, existing work is 
concentrated mainly on English, German, or multilingual datasets, while Hungarian 
remains underexplored for document-level reflection classification 
\citep{wulff2023utilizing, zhang2024classification, nehyba2023applications}.
This gap is important because Hungarian is a morphologically rich language, 
and Hungarian NLP benefits from language-specific resources and 
model development \citep{Nemeskey:2021a, yang-puli}.

This study investigates automatic reflection level classification in Hungarian student essays 
using two model families: (1) classical machine learning with TF-IDF and semantic 
embeddings, and (2) Hungarian transformer models fine-tuned for document classification. 
Because our dataset is strongly imbalanced, balancing strategies were evaluated 
systematically, and an ablation study was conducted to quantify the contribution 
of each modeling choice.

\paragraph{Our contributions} are the following:
\begin{enumerate}
    \item A comprehensive Hungarian analysis of automatic reflection-level classification in student essays, using a large expert-annotated corpus collected across multiple academic years.
    \item Comparing two modeling paradigms for Hungarian reflective text classification: classical machine learning with engineered features and fine-tuned Hungarian transformer models.
    \item Systematic evaluation of multiple imbalance-handling techniques to improve minority-class performance.
    \item An ablation study that clarifies the effect of model family and balancing strategy on classification performance.
\end{enumerate}

The rest of the paper will introduce the related works in Section~\ref{sec:literature_review}, followed by the methodology in 
Section~\ref{sec:methodology}. Experimental results are presented in Section~\ref{sec:results}, 
followed by a discussion in Section~\ref{sec:discussion}, and conclusions in Section~\ref{sec:conclusion}. 

\section{Related Work}
\label{sec:literature_review}

This section reviews relevant literature in two main areas.
 Firstly, it examines the previous literature about the 
 importance of reflective thinking in the field of 
 education. Secondly, it explores, the current state 
 of the automated reflective text classification systems, 
 highlighting key methodologies and findings from 
 prior research.

\subsection{Reflective thinking in education}

Multiple international frameworks such as the Tuning project
\citep{tuningacademy}, the OECD Learning Compass 2030 \citep{oecd} and
the EU Key Competences \citep{europeanconuncil} identify reflective
thinking or critical reflection as a generic, transformative or
transversal competency, respectively, and require educational
programmes to incorporate it into learning outcomes.

Reflective thinking is the deliberate process of analyzing and
evaluating one's own thoughts, experiences, and actions to gain deeper
insights and improve future decision-making. In university settings,
reflective thinking is often explored in the context of learning, for
example, by reviewing one's own understanding of the material
presented in theoretical courses and critically comparing it with
personal beliefs or practical experiences. Reflection also enables
students to interpret their experiences during practical placements,
fostering deeper learning and professional development. In many
professional fields, reflection is employed during the practical
training phase to support the development of students' professional
skills and the formation of their professional identity. This is
evident in education, in teacher training \citep{lee2005understanding,
  kaplar2020use, korthagen2022power}, or in early childhood educator
programs \citep{sumsion2000facilitating, cherrington2018early}, as
well as in medical education \citep{lim2023systematic, wald2010beyond,
  wald2019grappling}, nursing programs \citep{tashiro2013concept,
  newton2000uncovering}, and social work training
\citep{occhiuto2024reflecting}.

In professional training settings, as well as in the context of
lifelong learning, reflection is viewed as a means of continual
personal and professional development because it enables the
reflective thinker to withstand habitual action and to seek solutions
to problems which may be weakly defined or ambiguous
({ullmann2019automated}\cite{ullmann2019automated}) and arise during
human encounters.

Reflective writing is a pedagogical practice that can foster
reflective thinking, for example through journaling. Its summarizing
end product, the reflective essay, can serve as a record of the
reflective thinking process and can therefore be formally assessed by
teachers, using for instance rubrics.

\subsection{Automated reflective text classification systems}

Automated text classification systems have been widely 
studied in various domains, including sentiment analysis \citep{pilicita2025sentiment, grimalt2024sentiment}, 
topic categorization 
\citep{schulman2018text, tang2015automatic}, 
and educational text analysis \citep{ferreira2019text}. 
In the context of reflective text classification, 
several studies have explored the use of machine 
learning techniques to classify reflective essays 
based on different aspects. 

\cite{ullmann2019automated} conducted a foundational study on the automated analysis of reflective writing using machine 
learning approaches. The research evaluated 76 student essays mainly from second- and third-year health, business, and 
engineering programs, annotated through a large-scale crowdsourcing process involving thousands of contributors. The 
study adopted a comprehensive reflection model using eight categories: one depth dimension (reflection vs. non-reflection) 
and seven breadth dimensions, including description of experience, feelings, personal belief, awareness of difficulties, 
perspective, lessons learned, and future intentions. Using unigram features, several supervised learning algorithms were 
tested, including Support Vector Machines, Random Forest, Naïve Bayes, and Neural Networks. The models achieved accuracies 
between 70\% and 90\% (Cohen's K = 0.53-0.85), approximately 10\% lower than manual annotation performance. This work 
demonstrated that reflective categories in student essays can be reliably identified through automated means, providing a 
scalable foundation for reflection analytics in educational contexts.

As a continuation of the previous publication, \cite{alrashidi2022evaluating} proposed an automated reflective 
writing analysis system to classify reflection indicators in computer science students' essays using machine learning 
and natural language processing. The study applied the Reflective Writing Framework (RWF), focusing on the seven breadth 
categories of reflection as in the previous publication excluding the depth dimension of reflection. Using 74 student 
essays comprising 1,113 annotated sentences, the authors extracted linguistic features based on n-grams and part-of-speech 
(PoS) n-grams and employed a Random Forest classifier for binary detection of each indicator. The approach achieved accuracies 
ranging from 75\% to 96\% (Cohen's K = 0.17-0.67), demonstrating that such linguistic representations can effectively capture 
reflection-related features and enable automated assessment of reflective depth in student writing.

\cite{alrashidi2024machine} investigated the automatic classification of reflection depth in STEM student writings, 
focusing on Computer Science (CS) education. Building on the seven reflection indicators defined by \cite{ullmann2019automated} - 
description of experience, understanding, feelings, reasoning, perspective, new learning, and future action - the study explored 
both the breadth and depth dimensions of reflection. Using a dataset of 1,200 annotated sentences from CS students, the authors 
extracted four types of linguistic features: n-grams, part-of-speech (POS) n-grams, sentiment features (from AFINN, Bing, and 
NRC lexicons), and WordNet-Affect features. They experimented with a wide range of machine learning algorithms, including Random 
Forest, Support Vector Machine (SVM), Naïve Bayes, Bayesian Network, Logistic Regression, Neural Network, and Deep Learning models. 
A comparative evaluation also included the transformer model XLM-RoBERTa 
\citep{conneau2020unsupervisedcrosslingualrepresentationlearning} for cross-lingual reflection detection. 
Their proposed two-stage framework-first classifying reflection indicators, then predicting the depth of reflection 
(non-reflective, reflective, critically reflective) - achieved higher performance (Cohen's K = 0.70, accuracy = 82.85\%) 
than single-stage feature-based approaches. This work demonstrated that using breadth indicators as features significantly 
improves the automated detection of reflection depth and represents one of the first attempts to automate reflective writing 
assessment within CS education using machine learning and natural language processing (NLP).

\cite{solopova2023papagai} presents PapagAI, one of the first open-source systems for automated feedback on reflective essays, 
implemented as a hybrid AI approach combining symbolic and neural methods. The system processes student reflections 
(drawn from a German reflective corpus) and extracts a rich set of signals: it classifies emotional content at the 
sentence level using a fine-tuned RoBERTa \citep{liu2019robertarobustlyoptimizedbert}; it recognizes which Gibbs 
reflective cycle phase (description, feelings, evaluation, analysis, conclusion, or action plan) a given sentence belongs 
to via an ELECTRA \citep{clark2020electrapretrainingtextencoders} multi-class classifier; it assesses the overall 
“reflective level” of the text (e.g. descriptive, dialogical, transformative) using a BERT large model; it also 
applies BERTopic \citep{grootendorst2022bertopicneuraltopicmodeling} on sentence embeddings for topic modeling; 
and it derives linguistic features (e.g. sentence length, subordinate clause counts, discourse markers) via spaCy 
and RFTagger \citep{schmid-laws-2008-estimation}. These multiple signals are combined via a rule-based reasoner 
that selects from predefined feedback templates (prompts) to produce tailored formative comments. The authors 
contrast PapagAI with generative LLM approaches (such as GPT-style models), arguing that the hybrid design offers 
advantages in transparency, controllability, and alignment with didactic theory, while acknowledging tradeoffs in 
flexibility and expressiveness.

\cite{zhang2024classification} present a comprehensive comparison of shallow \seqsplit{machine-learning} models and pre-trained 
transformer models for document-level classification of reflective writing, using a dataset of 1,043 reflections 
collected from a German teacher-education program. The authors evaluate \seqsplit{sparse/textual} representations (bag-of-words 
and term frequency-inverse document frequency as TF-IDF) and psycholinguistic features derived from \cite{liwc2015}, 
feeding these features into several classical classifiers (e.g., Ridge, XGBoost, SGD). They contrast these baselines 
with fine-tuned transformer families BERT, RoBERTa and long-context transformers like
Longformer \citep{beltagy2020longformerlongdocumenttransformer} and BigBird \citep{zaheer2021bigbirdtransformerslonger}. 
Their results show a clear advantage for models that can process extended contexts: Longformer and BigBird achieved 
the highest classification accuracies ( $\approx$ 77\%), while shallow models remained below $\approx$ 60\% accuracy. 
The study therefore highlights the importance of document-level modeling and long-context capacity when automatically 
assessing reflective texts, and it underscores how modern transformer architectures improve the fidelity of automated 
reflection classification compared to traditional sparse or psycholinguistic feature approaches.

\cite{wulff2023utilizing} investigate the use of segment-level classification to automatically detect reflective categories 
in preservice physics teachers' writings, using a fine-tuned BERT model. Their corpus has 270 reflections from 92 preservice 
teachers, each divided into segments that express a single cohesive idea. Drawing on reflection support models e.g. 
\citep{korthagen2005levels, ullmann2019automated}, the authors label units into five canonical reflective elements: 
circumstances, description, evaluation, alternative devising, and derivation of consequences. In addition to BERT, 
they implement feedforward neural networks and LSTM baselines, and compare with ELMo 
\citep{peters2018deepcontextualizedwordrepresentations} + SVM and simpler count-based representations. 
Their models are carefully tuned via grid search and evaluated under sensitivity analyses 
(sampling fractions of training data). They apply Integrated Gradients via Captum \citep{captum} to highlight token 
contributions in classification. The results indicate that BERT significantly outperforms the baselines, and that 
interpretability methods can enhance the pedagogical transparency of automated reflection classification.

\cite{nehyba2023applications} present Applications of Deep Language Models for Reflective Writings, a 
robust study of reflective writing analysis in teacher education contexts, employing deep pretrained models 
in a multilingual setting. They detail a strict annotation process of student-teachers' reflective journal 
sentences, and experiment with both traditional machine-learning representations (e.g., bag-of-words) and 
deep models (notably XLM-Roberta). Their results indicate strong performance (accuracy from $\approx$76.6\% 
up to $\approx$100\% in high-confidence subsets) and suggest the approach is applicable across more than 100 
languages with minimal loss in accuracy. The study underscores the practical value of pretrained architectures 
for capturing reflective thinking in texts, emphasises annotation and multilingual scalability, and provides an 
open-source resource framework. The present work builds on this by exploring a Hungarian-language domain and 
investigating whether a language-specific pretrained model can outperform or complement a multilingual model in 
reflective-writing classification.

\cite{chong2020analysis} explore the use of natural-language-processing (NLP) 
and fuzzy logic for analysing students' reflective writing in a psychology course. Their 
corpus comprises 47 reflective journals, collected via end-of-semester surveys, and pre-processed 
using NLTK \citep{nltk} for cleaning and 
tokenisation. The authors build a domain-specific keyword database and apply a fuzzy logic system 
that uses count-based features (keyword counts, sentiment-oriented terms) to categorise each journal 
at a document level into predefined reflection-levels. Their analysis finds that the majority of 
student writings fall into lower levels of reflection (levels 1-2). Although the methodology is 
relatively lightweight compared to more recent deep-learning-based approaches, this work demonstrates 
the value of combining domain lexica and rule-based logic in automated reflection analytics. It thus 
provides an early proof-of-concept in this space, and serves as a useful baseline when comparing more 
sophisticated ML/NLP pipelines for reflective writing classification.

\section{Methodology}
\label{sec:methodology}

\subsection{Dataset}

\begin{figure}
    \centering
    \includegraphics[width=0.9\columnwidth]{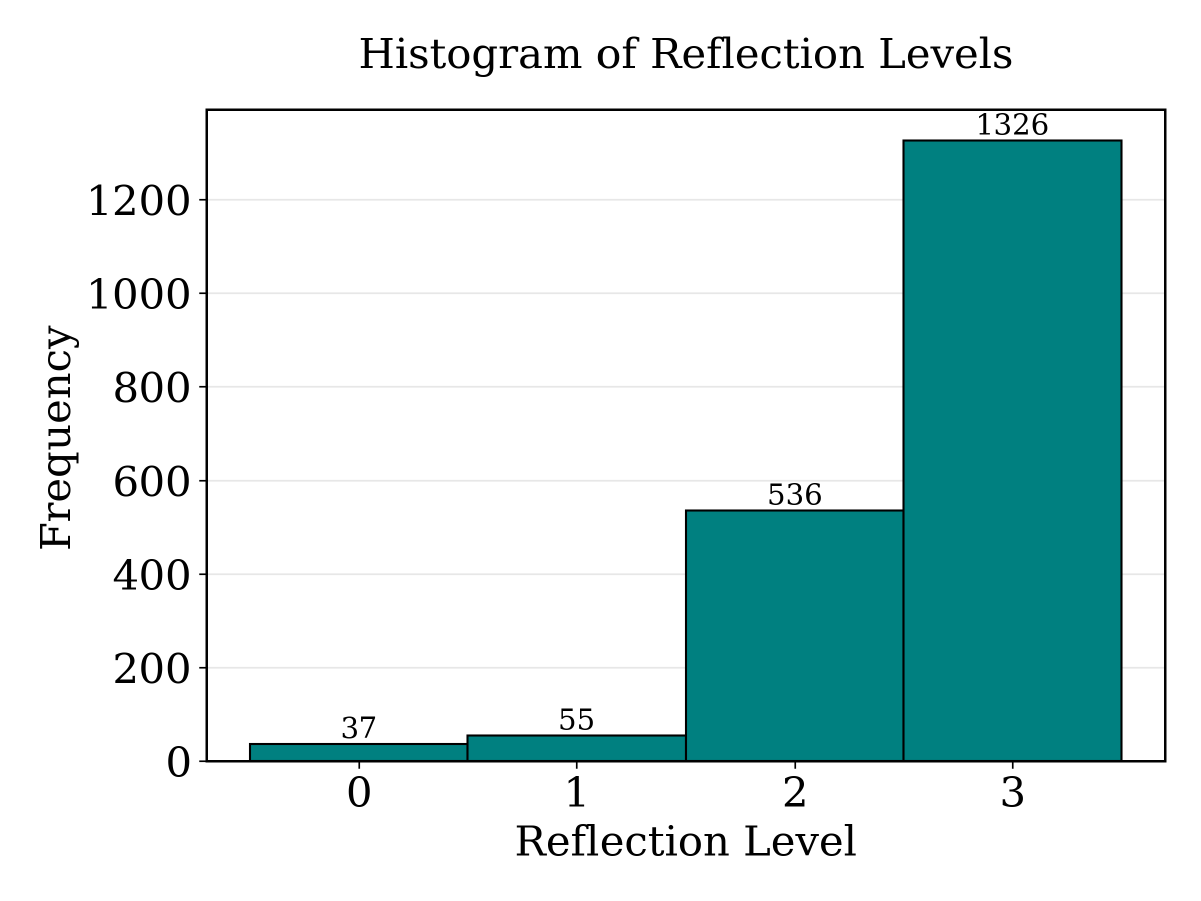}
    \caption{Distribution of reflection levels in the dataset. From 0 (no reflection) to 3 (high reflection), where 1.8\% of the essays are rated as level 0, 2.8\% as level 1, 27.4\% as level 2 and 68\% as level 3.}
    \label{fig:reflection_histogram}
\end{figure}

The dataset used in this study was collected by experts in the field
of education and reflective writing at a large public University in
Central Europe. They collected essays in Hungarian from students who
completed their studies over a period of 4 consecutive years. Throughout
the six semesters of the Early Childhood Education programme, students
complete pedagogical practice placements in nurseries, where they
receive field-based training. At the end of each semester, they are
required to write a reflective essay addressing guiding questions
related to their practical experiences. To support the preparation of
these essays, students receive detailed written guidance (based on the
work of \cite{gyongykinga}), including an assessment rubric that can
be used for self-evaluation prior to submitting the final work.

Experts rated the essays on multiple dimensions (formal, structural, and content-related), 
including reflection level, format, and text creation. Only reflection-level labels are 
used in this study. The dataset contains almost 1954 annotated essays from roughly
450 students. The dataset is not publicly available due to privacy and ethical considerations.

The distribution of the reflection levels in the dataset is heavily imbalanced,
as demonstrated in Figure~\ref{fig:reflection_histogram}. The reflection levels are
rated on a scale from 0 to 3 where 0 means no reflection, 1 means low
reflection, 2 means medium reflection and 3 means high reflection.

Ethical permission for this research was granted by the ethics committee of Anonymous University(Number).

\subsection{Data Preprocessing}

The essays from the dataset went through an extensive preprocessing phase to
make them suitable for the text processing algorithms (Shallow Machine Learning,
Transformer models).

Firstly, the essays came in different formats (docx, pdf, txt) and the text had
to be extracted from these documents. The text extraction was done using the
\texttt{python-docx} \citep{python_docx} Python Library for docx files and
\texttt{PdfPlumber} \citep{pdfplumber} Python library was used for pdf files\footnote{PyPDF2 \citep{pypdf2} was also tried, but it had issues with Hungarian characters and inserted random spaces inside words.}.

\begin{figure}
    \centering
    \includegraphics[width=0.9\columnwidth]{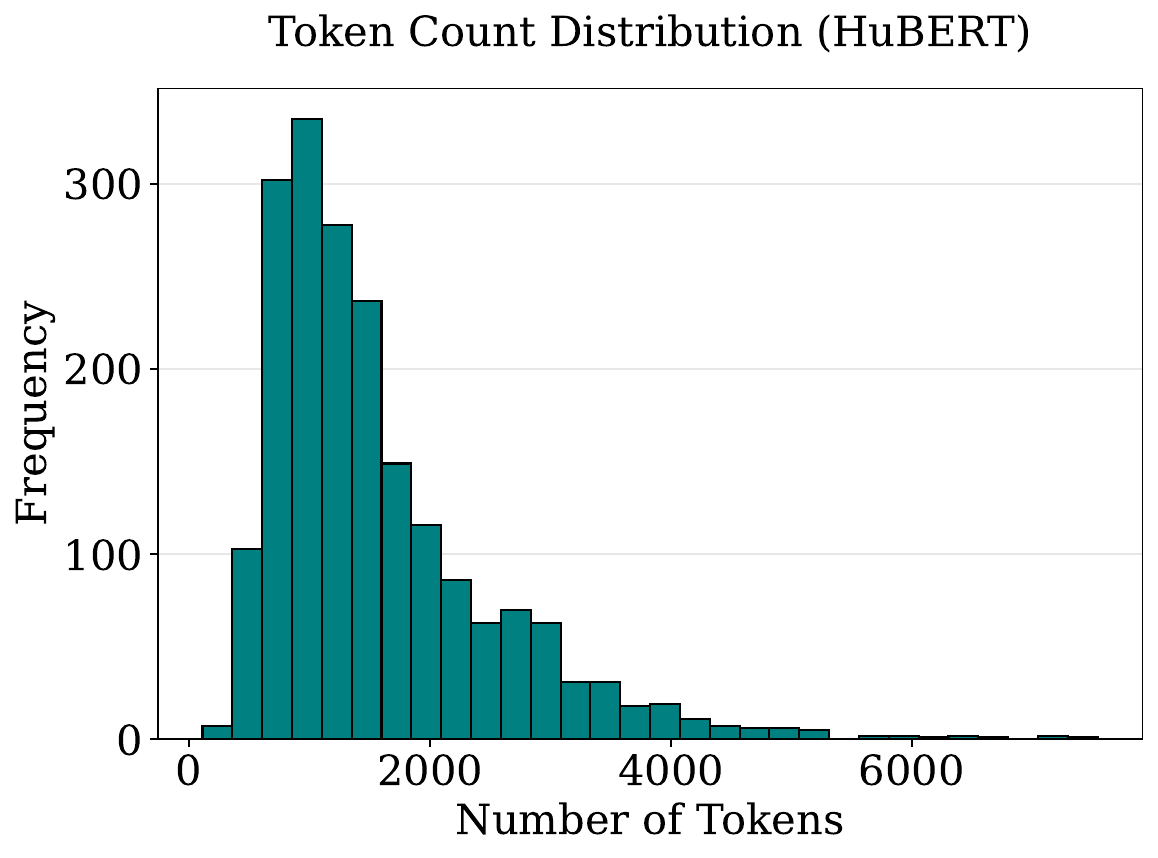}
    \caption{Token count distribution generated by the HuBERT tokenizer. The histogram shows a right-skewed distribution, with most samples containing fewer than 2,000 tokens and peaks around 1,000 tokens.}
    \label{fig:token_histogram}
\end{figure}

Additionally, the documents contained some cover pages as template which are not
relevant for the task, so these were removed from the text manually. Also, personally identifiable information was present in the essays, such as names, places,
organizations, dates, etc. These were removed with a three-fold approach: first
using \texttt{huspacy} \citep{huspacy} python
library with their best model possible, namely \texttt{hu\_core\_news\_trf\_xl}
a model built on XLM-RoBERTa-large architecture  for named entity
recognition removing most of the names, organisations and places and replace
them with a generic placeholder like \texttt{[PER]}, \texttt{[LOC]},
\texttt{[ORG]}. In the second step, the text was cleaned up by removing whitespaces, new
lines, tabs, unnecessary punctuation marks and spaces. Finally, the text was
cleaned using a custom regular expression (for more details see Additional File 3) to remove dates, month names,
roman numerals, and specific Hungarian words which could contain locality
information (for example word "kerület" which means district in Hungarian and it
was not covered by the named entity recognition).

As a last step, the texts were manually checked again for any remaining
sensitive information and saved into a dataset format called \texttt{.arrow} \citep{apache_arrow}
which is a columnar format that is suitable for large datasets and efficient for
processing with machine learning libraries like HuggingFace's \texttt{datasets}.

\subsection{Shallow Machine Learning Models}
\label{sec:shallow_models}

The most commonly used models in the literature for reflection detection are the
following: Random Forest (RF), XGBoost, CatBoost and RidgeClassifier. These models were tested
on the Hungarian essay dataset 
used in this study. The features for these models in every case were selected from different
sources:
\begin{itemize}
    \item TFIDF features: using the \texttt{TFIDFVectorizer} from the
    \texttt{sklearn} Python library, which calculates the term frequency-inverse
    document frequency of the words in the essays, makes a constant length
    vector representation of the essays, which can be used by the models.
    \item Document embeddings: using pre-trained document embeddings to represent the
    essays. Based on the Massive Text Embeddings benchmark one of the best model
    in the multi-lingual category is the \texttt{\seqsplit{Qwen3-4B-Embedding} by \cite{zhang2025qwen3embeddingadvancingtext}} model.
    Qwen3 embeddings gives a fixed-length 2560-dimensional vector representation
    of the essays, which can be used by the models.

\end{itemize}

The models were trained using the \texttt{sklearn} library and the training was
done with a 5-fold cross-validation in each case using \texttt{StratifiedKFold}. The random seed for all
experiments was set to 42.

The data was split into 80\% training and 20\% testing split on the essay level, and the results
were evaluated using the following metrics:
\texttt{Accuracy, Precision, Recall, F1-score, ROC-AUC}. The results were visualized using 
\texttt{Confusion Matrices} and \texttt{ROC Curves}.

As outlined in Figure~\ref{fig:reflection_histogram}, the Hungarian essays dataset
is heavily imbalanced. Firstly the models were trained on the original data
without any balancing techniques to obtain the baseline results.

After that multiple balancing techniques were evaluated:
\begin{itemize}
    \item Class weighting: using the \texttt{class\_weight} parameter in the
    models to give more weight to the minority classes and less weight to the
    majority classes. XGBoost uses sample weights instead of the class weights.
    \item Oversampling techniques: using multiple oversampling techniques
    \texttt{\seqsplit{SMOTE}} \citep{Chawla_2002} (Synthetic Minority Over-sampling Technique) and
    \texttt{\seqsplit{ADASYN}} \citep{4633969} (Adaptive Synthetic Sampling), \texttt{\seqsplit{Random Over Sampling}} \citep{japkowicz2002class} and,
    \texttt{\seqsplit{BorderlineSMOTE}} \citep{han2005borderline} from the \texttt{\seqsplit{smote-variants}} \citep{smote_variants} Python library to
    generate synthetic samples from the minority classes.
    \item Ensemble techniques with oversamplers: using
    \texttt{\seqsplit{EasyEnsembleClassifier}} \citep{liu2008exploratory}, \texttt{\seqsplit{SMOTEBoostClassifier}} \citep{chawla2003smoteboost},
    \texttt{\seqsplit{SMOTEBaggingClassifier}} \citep{wang2009diversity}, \texttt{\seqsplit{AdaUBoostClassifier}} \citep{sun2007cost},  and
    \texttt{\seqsplit{CompatibleAdaBoostClassifier}} \citep{beja2023overview} from the \texttt{\seqsplit{imbalanced-ensemble}} \citep{imbalanced_ensemble}
    Python library, to combine the oversampling techniques with ensemble
    methods.
\end{itemize}

All the results coming from these models were K-fold cross-validated with 5
folds, and the results were averaged over the folds.

\subsection{Transformer Models}
\label{sec:transformer_models}

Transformer models are powerful deep learning models that have revolutionized
the field of natural language processing (NLP) in recent years. The training of
transformer models in Hungarian language is challenging due to the limited
availability of large-scale annotated datasets and pre-trained models
specifically designed for Hungarian. However, several transformer models have
been developed that support Hungarian language. From these models, the following
were used for this paper:
\begin{itemize}
    \item \texttt{SZTAKI-HLT/hubert-base-cc}: \cite{Nemeskey:2021a} introduced a BERT-based model trained on a
    Hungarian subset of the Common Crawl and a snapshot of the Hungarian
    Wikipedia.
    \item \texttt{NYTK/PULI-BERT-Large}: The work of \cite{yang-puli}, they introduced a Hungarian BERT large model based on
    MegatronBERT.
\end{itemize}

Both hubert and PULI-BERT-Large have a short context window (512 tokens),
which is a limitation for document level classification tasks.
To address this limitation, chunking with overlap was applied
(128 token for chunking and 64 token for overlap which gives a good 
performance-efficiency as shown by \cite{jaiswal2023breaking}) 
to split the essays into smaller chunks
and then aggregating the chunk level predictions using average pooling.

These models were fine-tuned on the Hungarian essay dataset for the task of
reflection level classification. For each model, the pre-trained base model was
loaded from the HuggingFace model hub using the \texttt{transformers} Python
library. Then a classification head was added on top of the base model to
perform document level reflection classification on the 4 categories. The base
model weights were frozen in some cases to reduce the number of trainable
parameters or not frozen to allow the model to learn the specific task better.

To handle data imbalance during the training of the transformer models, multiple
techniques were evaluated:
\begin{itemize}
    \item Oversampling: Random oversampling was used to make more samples for
    the minority classes. To make the new samples by oversampling, the
    RandomOverSampler (ROS) from the \texttt{imblearn} Python library was used.
    \item Text Augmentation: Using text augmentation techniques like
    backtranslation to generate similar samples for the minority classes. During
    backtranslation only those samples were considered which had at least 80\%
    embedding similarity with the original text to avoid generating too
    different samples. To make the backtranslated data, English was used as the
    pivot language. For the translation to English and back to Hungarian the
    \texttt{Helsinki-NLP/opus-mt-hu-en} \citep{tiedemann2023democratizingneuralmachinetranslation} 
    and \texttt{Helsinki-NLP/opus-mt-en-hu} \citep{tiedemann2023democratizingneuralmachinetranslation}
    models from HuggingFace were used. For calculating the embedding similarity
    the \texttt{\seqsplit{sentence-transformers/paraphrase-multilingual-MiniLM-L12-v2}} \citep{reimers-2019-sentence-bert}
    model was used.
    \item Loss Functions: Besides the standard Cross Entropy Loss, Weighted
    Cross Entropy Loss, Focal Loss and Dice loss were also tried to give more
    importance to the minority classes during training.
\end{itemize}

The training pipeline was implemented using the \texttt{\seqsplit{transformers}} \citep{huggingface_transformers} and
\texttt{datasets} \citep{huggingface_datasets} Python libraries from HuggingFace. 
Early stopping was used based on the \textbf{validation accuracy}
and patience of 5 epochs allowing the model to stop training early and prevent overfitting. During training linear
scheduler, AdamW optimizer were used with weight decay and the random seed was 42. The experiments were
run on a single NVIDIA 3090 GPU with 24GB of VRAM.

As the shallow machine learning models, the transformer models were also trained
using an 80\% training and 20\% testing split, and the results were calculated
with stratified 5-fold cross/validation using the same metrics as described in
\ref{sec:shallow_models}.

\begin{figure*}
  \centering
  \begin{subfigure}[b]{0.48\columnwidth}
    \includegraphics[width=0.915\textwidth]{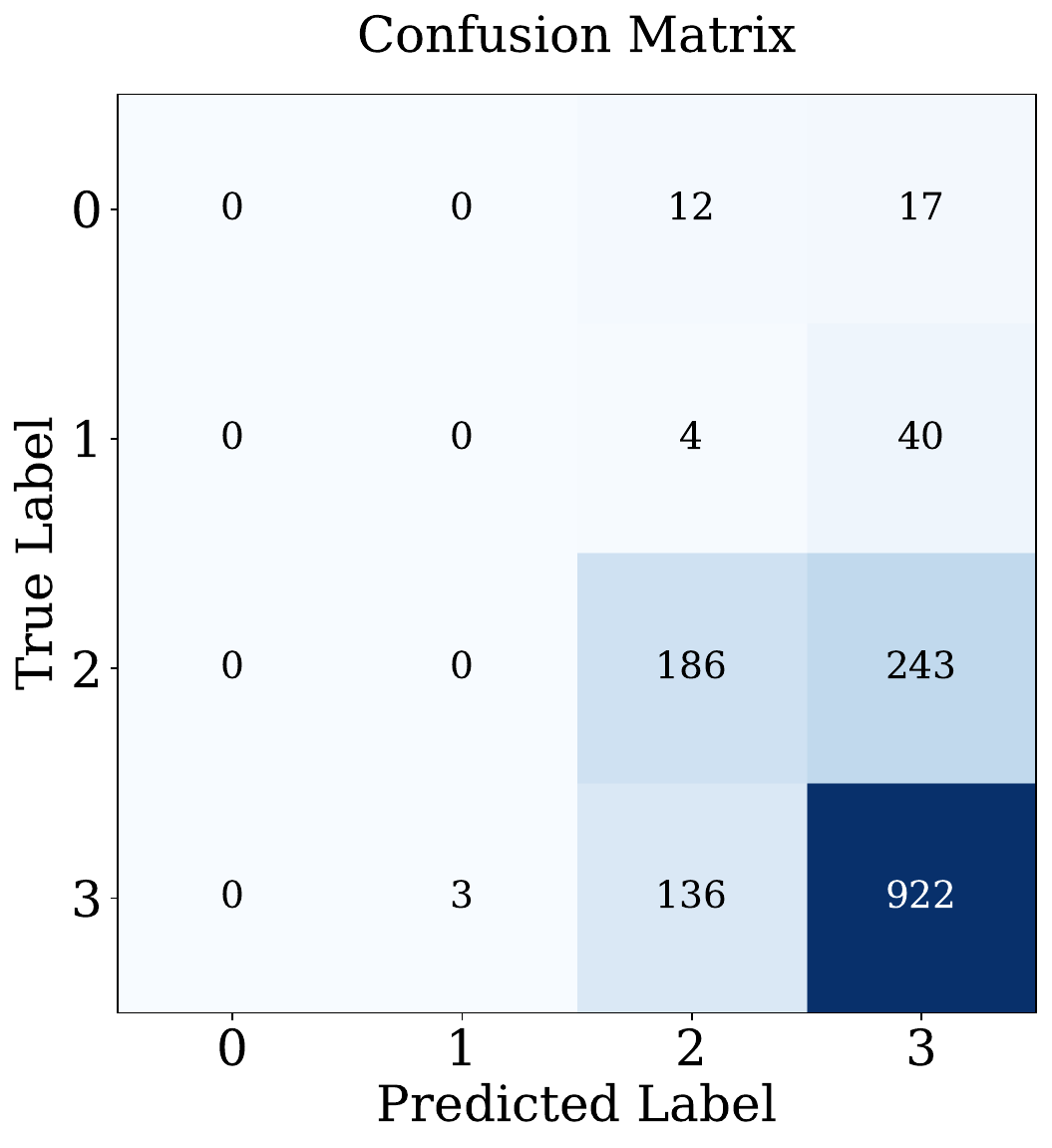}
    \caption{}
    \label{fig:confusion_matrix_shallow}
  \end{subfigure}
  \hfill
  \begin{subfigure}[b]{0.48\columnwidth}
    \includegraphics[width=0.99\textwidth]{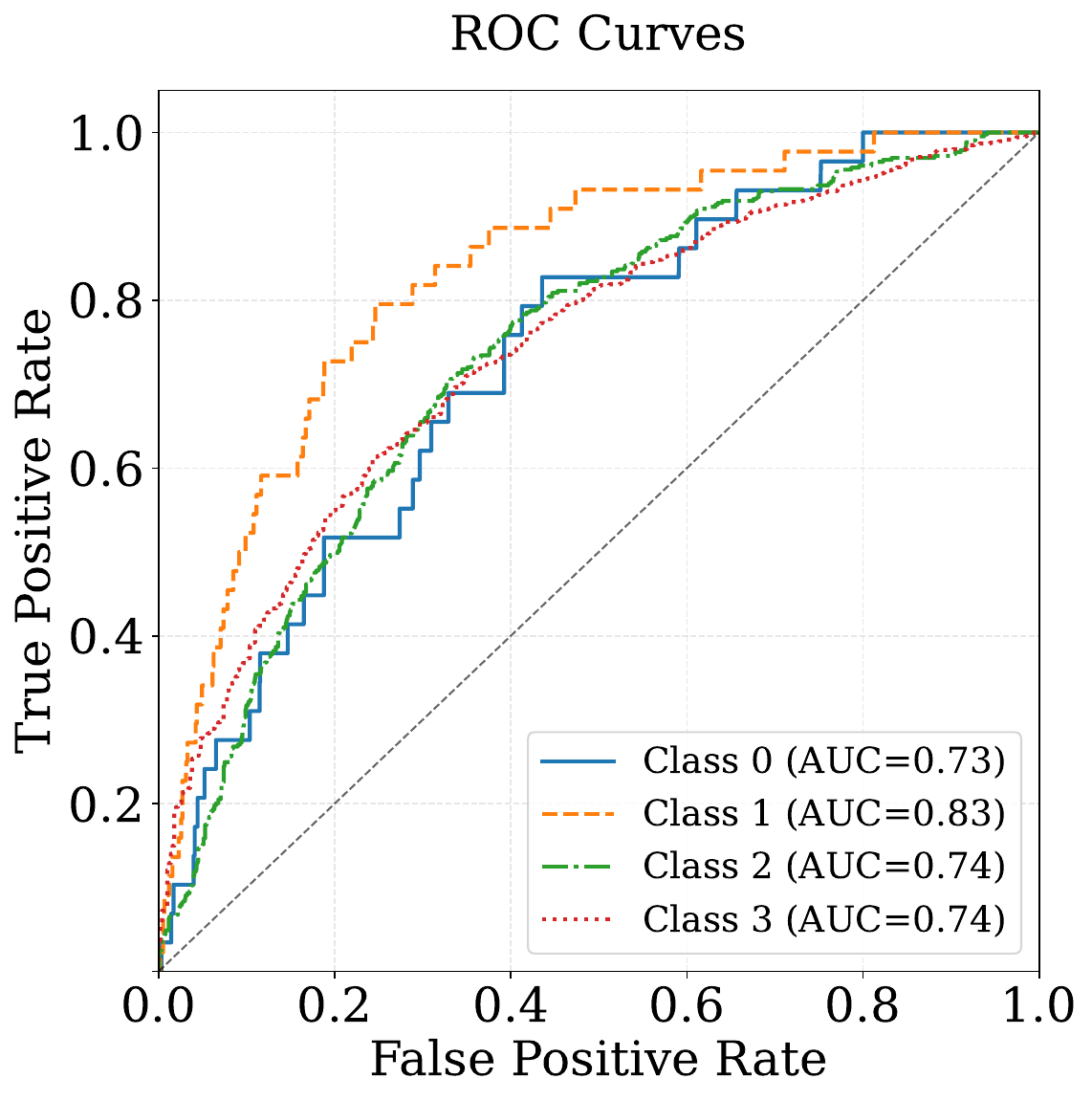}
    \caption{}
    \label{fig:roc_curves_shallow}
  \end{subfigure}
  \caption{Confusion matrix (a) and ROC curves (b) of the best Shallow Machine Learning model (SMOTEBoost + Qwen3 embeddings + Decision tree with 12 maximum depth) on the Hungarian essays dataset. The results are averaged over 5 folds.}
  \label{fig:combined}
\end{figure*}

\section{Experimental Results}
\label{sec:results}

In this section the results of the experiments are presented. First the the
results of the Shallow Machine Learning models are described then finally the
results of the Transformer models.

\subsection{Shallow Machine Learning models}

\begin{table*}[!t]
    \centering
    \resizebox{1.0\textwidth}{!}{%
    \begin{tabular}{|c|c|c|c|c|c|c|}
    \hline
        \textbf{Model} & \textbf{OS.} & \textbf{Qwen3-E.} & \textbf{Accruacy} & \textbf{F1} & \textbf{ROC Auc} & \textbf{Overall} \\ \hline
        \textbf{EasyEnsemble} & \cmark & \cmark & \num{0,6320} & \num{0,6481} & \num{0,7240} & \num{0,6680} \\ \hline
        \textbf{AdaUBoost} & \cmark & \cmark & \num{0,6967} & \num{0,6197} & \num{0,7426} & \num{0,6863} \\ \hline
        \textbf{XG-Boost} & \xmark & \cmark & \num{0,7082} & \num{0,6668} & \num{0,7178} & \num{0,6976} \\ \hline
        \textbf{CompatibleAdaBoost} & \cmark & \cmark & \num{0,7075} & \num{0,6568} & \num{0,7364} & \num{0,7003} \\ \hline
        \textbf{Random Forest} & \xmark & \cmark & \num{0,7056} & \num{0,6613} & \num{0,7360} & \num{0,7010} \\ \hline
        \textbf{CatBoost} & \xmark & \cmark & \num{0,7012} & \num{0,6540} & \num{0,7480} & \num{0,7010} \\ \hline
        \textbf{SMOTEBagging} & \cmark & \cmark & \num{0,6973} & \num{0,6623} & \num{0,7569} & \num{0,7055} \\ \hline
        \textbf{RidgeClassifier} & \xmark & \cmark & \textbf{\num[mode=text]{0,7223}} & \textbf{\num{0,6833}} & \num{0,7339} & \num{0,7131} \\ \hline
        \textbf{SMOTEBoost} & \cmark & \cmark & \num{0,7088} & \num{0,6808} & \textbf{\num{0,7633}} & \textbf{\num{0,7176}} \\ \hline
    \end{tabular}
    }
    \vspace{0.2cm}
    \caption[]{Best result of each model using the Hungarian 
    essay dataset. The results are averaged over 5 folds. The best 
    results in each column are in bold. Green checkmark 
    (\cmark) indicates the usage of the specific technique, 
    while red cross (\xmark) indicates its absence. The abbreviation OS.
    stands for Oversampling, while Qwen3-E. stands for Qwen3 Embeddings.}
    \label{tab:shallow_results}
\end{table*}

As presented in \ref{sec:shallow_models}, multiple Shallow Machine Learning
models  with multiple different imbalanced data handling techniques were trained
and evaluated on the training dataset. The results of the models are presented
in Table~\ref{tab:shallow_results}. This is not the complete table (for full table see Additional File 1 from the supplementary), it only indicates
the best result of each model with the best hyperparameters (like oversampling
technique, usage of semantic embeddings etc.).

As summarized in Table~\ref{tab:shallow_results}, the best results were achieved by
the SmoteBoostClassifier which are using the SMOTE oversampling  model, as a base estimator it uses a Decision tree with 12 maximum depth and taking the Qwen3 semantic embeddings. This model achieved an average score of
\textbf{0.7176} (averaged over the 3 metrics), with \num{0,7089}  mean accuracy (with \num{0,0263} standard deviation), \num{0,6808} mean F1-score
(with \num{0,0309} standard deviation)
and \num{0,7633} mean ROC-AUC score (with \num{0,0285} standard deviation) averaged over the 5 folds.

The confusion matrix of the best model (SMOTEBoost + Qwen3 embeddings + Decision tree with 12 maximum depth) is
shown in Fig.~\ref{fig:confusion_matrix_shallow}. The confusion matrix indicates higher performance on majority 
classes, while minority-class samples are frequently misclassified as majority-class samples.
Even misclassified 3 samples from the majority class 3 as minority class 1.

The ROC curve for each class with the best Shallow Machine Learning model is
shown in Fig.~\ref{fig:roc_curves_shallow}. The ROC curves show that the SMOTEBoost model
can distinguish in a balanced way between the classes, as the AUC scores of each class are above
0.73 with a better performance on the class 1.

In general, classical machine learning models which are able to handle imbalanced
data well, like XGBoost, RidgeClassifier, Random Forest achieved good results on
this dataset. However these models mostly learnt to classify the majority class
well, not even predicting the minority classes correctly as shown on the Fig.~\ref{fig:confusion_matrix_shallow}. The introduction of
oversampling/undersampling techniques and class balancing made the models
classify the minority classes occasionally, but not necessarily improve the overall
performance of the models. 
Besides the EasyEnsemble model, the introduction of ensemble techniques with oversampling/undersampling, made similar 
or even better results than the base shallow machine learning models.

\begin{figure*}[t!]
  \centering
  \begin{subfigure}[b]{0.48\textwidth}
    \includegraphics[width=0.9\textwidth]{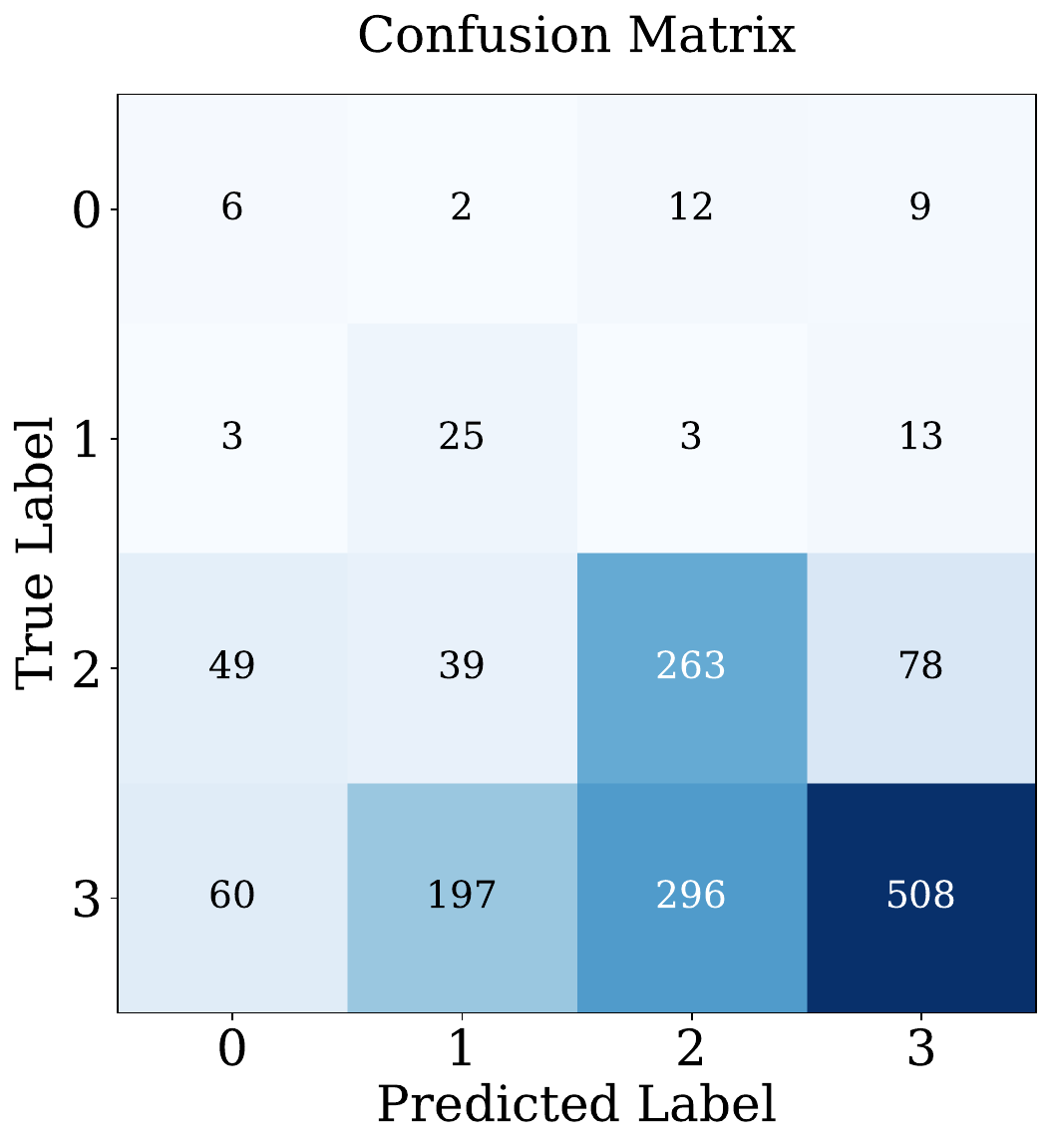}
    \caption{}
    \label{fig:bad_confusion_matrix_shallow}
  \end{subfigure}
  \hfill
  \begin{subfigure}[b]{0.48\textwidth}
    \includegraphics[width=0.95\textwidth]{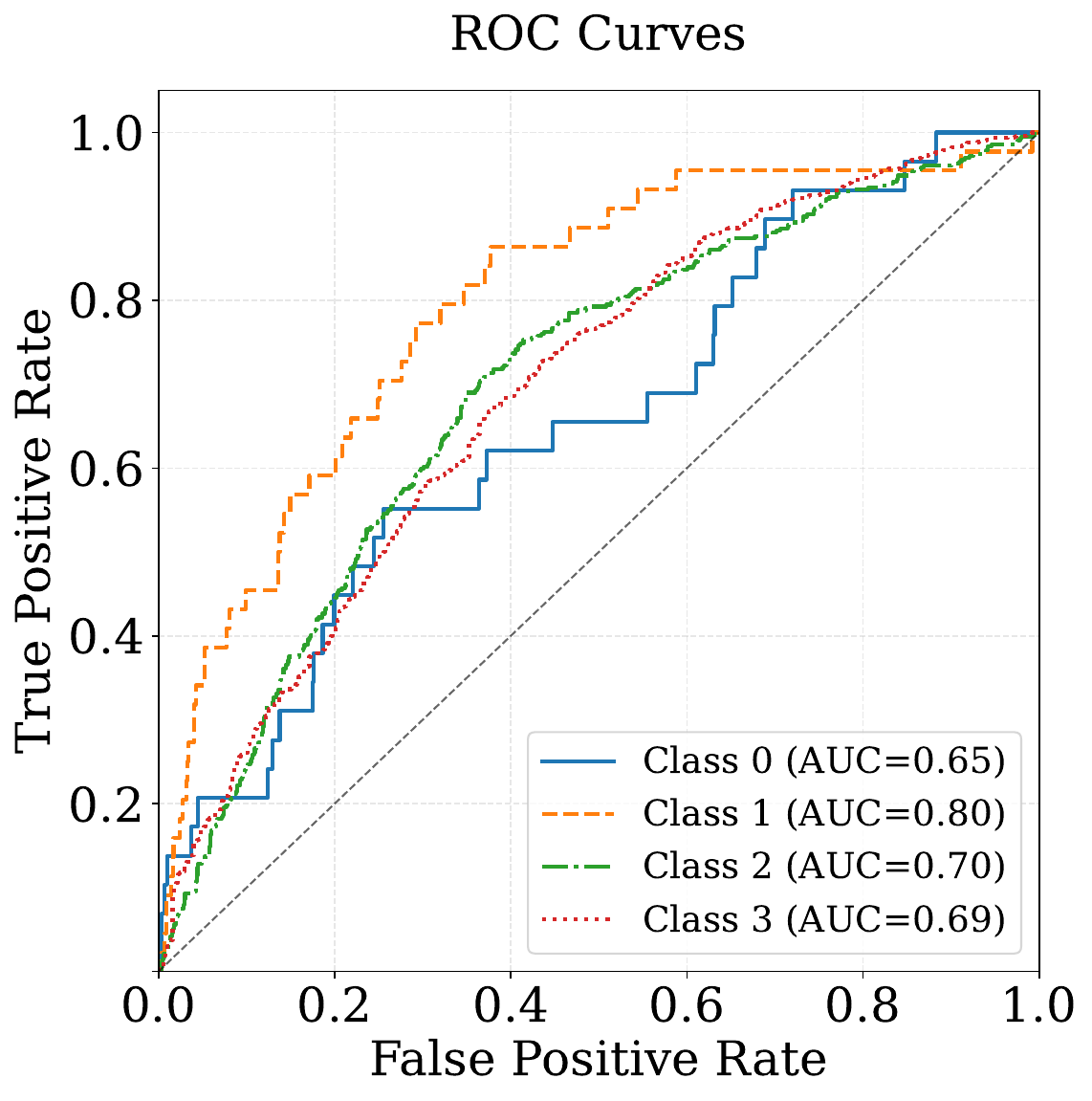}
    \caption{}
    \label{fig:bad_roc_curves_shallow}
  \end{subfigure}
  \caption{Confusion matrix (a) and ROC curves (b) of an example where RidgeClassifier produces high ROC-AUC score but low accuracy on the hungarian essays dataset. The results are averaged over 5 folds.}
  \label{fig:combined_ridge}
\end{figure*}

Additionally, in some cases handling the imbalance made
the models worse because they started to misclassify the majority classes
heavily. One example when this happened is with the RidgeClassifier. It produced
a good ROC-AUC score, but does not necessarily mean high performance in the
other metrics as well like accuracy and F1-score. For example RidgeClassifier
using Qwen3 embeddings with alpha parameter of 5, no class balancing and \seqsplit{SMOTE} oversampling technique achieved \num{0,7131} ROC-AUC score (with \num{0,0379} standard deviation),
but only \num{0,5162} accuracy (with \num{0,0356} standard deviation) over the 5 fold. The ROC-AUC curves and confusion matrix of this model are
shown in Fig.~\ref{fig:bad_confusion_matrix_shallow} and Fig.~\ref{fig:bad_roc_curves_shallow}.

\begin{table*}[t!]
\centering
\resizebox{1.0\textwidth}{!}{%
\begin{tabular}{|c|cc|cc|cc|cc|}
\hline
\textbf{Method name} &
\multicolumn{2}{c|}{\textbf{Accuracy}} &
\multicolumn{2}{c|}{\textbf{F1}} &
\multicolumn{2}{c|}{\textbf{ROC AUC}} &
\multicolumn{2}{c|}{\textbf{Overall}} \\
 & \textbf{\cmark} & \textbf{\xmark} & \textbf{\cmark} & \textbf{\xmark} & \textbf{\cmark} & \textbf{\xmark} & \textbf{\cmark} & \textbf{\xmark} \\ \hline
Class balancing & \num{0,6408} & \textbf{\num{0,6672}} & \num{0,6283} & \textbf{\num{0,6404}} & \num{0,7083} & \textbf{\num{0,7134}} & \num{0,6591} & \textbf{\num{0,6736}} \\ 
Oversampling & \num{0,6481} & \textbf{\num{0,6724}} & \textbf{\num{0,6358}} & \num{0,6294} & \num{0,7087} & \textbf{\num{0,7176}} & \num{0,6604} & \textbf{\num{0,6731}} \\ 
Semantic Embeddings & \num{0,5977} & \textbf{\num{0,6259}} & \textbf{\num{0,5891}} & \num{0,5772} & \textbf{\num{0,6627}} & \num{0,6446} & \textbf{\num{0,6165}} & \num{0,6159} \\ \hline
\end{tabular}
}
\vspace{0.2cm}
\caption{Ablation study results of different techniques on the Hungarian essays dataset. The results are averaged over the models which were 5-fold cross-validated. Green checkmark (\cmark) indicates the usage of the specific technique, while red cross (\xmark) indicates its absence. The best results in each row are in bold.}
\label{tab:method_performance}
\end{table*}

In Table \ref{tab:method_performance} the results of 
an ablation study are presented. 
This shows the average performance 
of the models when using or not using a 
specific technique like class balancing, 
oversampling and semantic embeddings.
One finding is that the models without class 
balancing achieved better results in every metric 
compared to the models with class balancing. In this dataset, 
models without class balancing obtained higher average scores across these metrics
, as the models are able to learn the class distribution well without 
any additional techniques and class balancing made the 
models misclassify more samples from the majority class.

Similarly, oversampling techniques also did not necessarily 
improve the overall performance of the models. Oversampling is associated with higher 
F1, while non-oversampled settings show higher Accuracy and ROC-AUC.
The usage of semantic embeddings improved the ROC-AUC scores 
of the models significantly, however the accuracy and 
F1-scores were better without using semantic embeddings. 
This shows that semantic embeddings can help the models to 
distinguish between the classes better.
These techniques can help the models to classify the minority 
classes better, however they do not necessarily improve the 
overall performance of the models. 

Overall, the ablation study shows that the effectiveness 
of these techniques can vary depending on the specific 
dataset and model used. It is important to note that 
due to the high class imbalance in the dataset, the 
models without any balancing techniques are mostly 
biased towards the majority class, which can lead to 
higher accuracy but lower performance on minority classes. 
The usage of these techniques made the models to at least 
classify some samples from the minority classes correctly, 
which is important in imbalanced classification tasks.

% Another important finding is the usability of oversampling/undersampling based
% ensemble techniques which most of the times made less performant classifiers,
% due it classified minority classes better, but missclassified more samples from
% the majority class. This is because of the high class imbalance, as the majority
% class has 10 times more samples than the minority classes and because of the
% noise in the data.

% The introduction of transformer-based semantic embeddings also improved the
% general performance of almost every model by a small margin. Models like Random
% Forest with Qwen3 embeddings also achieved very good results, with an average
% score of 0.6920, which is better than most of the models using TFIDF features.
% Additionally the ROC-AUC score of the models using Qwen3 embeddings is also
% significantly better, which shows that the model is able to distinguish between
% the classes better.

% This shows the advantage of using semantic embeddings over TFIDF features, as
% the embeddings are able to capture the semantic meaning of the essays better,
% which helps the models to classify the essays better.

\subsection{Transformer models}

As mentioned in Section \ref{sec:transformer_models}, two Hungarian pre-trained
transformer models were supervised fine-tuned on the training dataset. The
best results of the selected models and selected hyperparameters (oversampling/data augmentation, different loss functions and backbone freezing) are presented in
Table~\ref{tab:transformer_results} (for more results see Additional File 2).

Generally the PULI-BERT-Large model achieved better results than the hubert-base-cc in the 
same hyperparameter settings. The best overall results were achieved by the PULI-BERT-Large 
model using Cross-Entropy loss function with backtranslation technique, 
achieving an average score of \textbf{\num{0,6872}} (averaged over the 3 metrics), 
with \num{0,7095} mean accuracy (with \num{0,0123} standard deviation), \num{0,6762}  mean F1-score 
(with \num{0,0244} standard deviation) and \num{0,6760} mean ROC-AUC score (with \num{0,0196} standard deviation).

\begin{table*}[!t]
    \centering
    \resizebox{1.0\textwidth}{!}{%
    \begin{tabular}{|c|c|c|}
    \hline
        \textbf{Model} & \textbf{hubert-base-cc} & \textbf{PULI-BERT-Large} \\ \hline
        \textbf{Loss function} & Focal & Cross-Entropy \\ \hline
        \textbf{Oversampling/Backtranslation} & \cmark & \cmark \\ \hline
        \textbf{Backbone freeze} & \xmark & \xmark \\ \hline
        \textbf{Accuracy} & \num{0,6967} & \textbf{\num{0,7095}} \\ \hline
        \textbf{F1} & \num{0,6601} & \textbf{\num{0,6762}} \\ \hline
        \textbf{ROC Auc} & \num{0,6962} & \textbf{\num{0,6760}} \\ \hline
        \textbf{Overall} & \num{0,6843} & \textbf{\num{0,6872}} \\ \hline
    \end{tabular}
    }
    \vspace{0.2cm}
    \caption[]{Best result of each Transformer model using the Hungarian essay dataset. The results are averaged over 5 folds. 
    The best results in each column are in bold. Green checkmark (\cmark) indicates the usage of the specific technique, 
    while red cross (\xmark) indicates its absence.}
    \label{tab:transformer_results}
\end{table*}

\begin{figure*}[t!]
  \centering
  \begin{subfigure}[b]{0.48\textwidth}
    \includegraphics[width=0.9\textwidth]{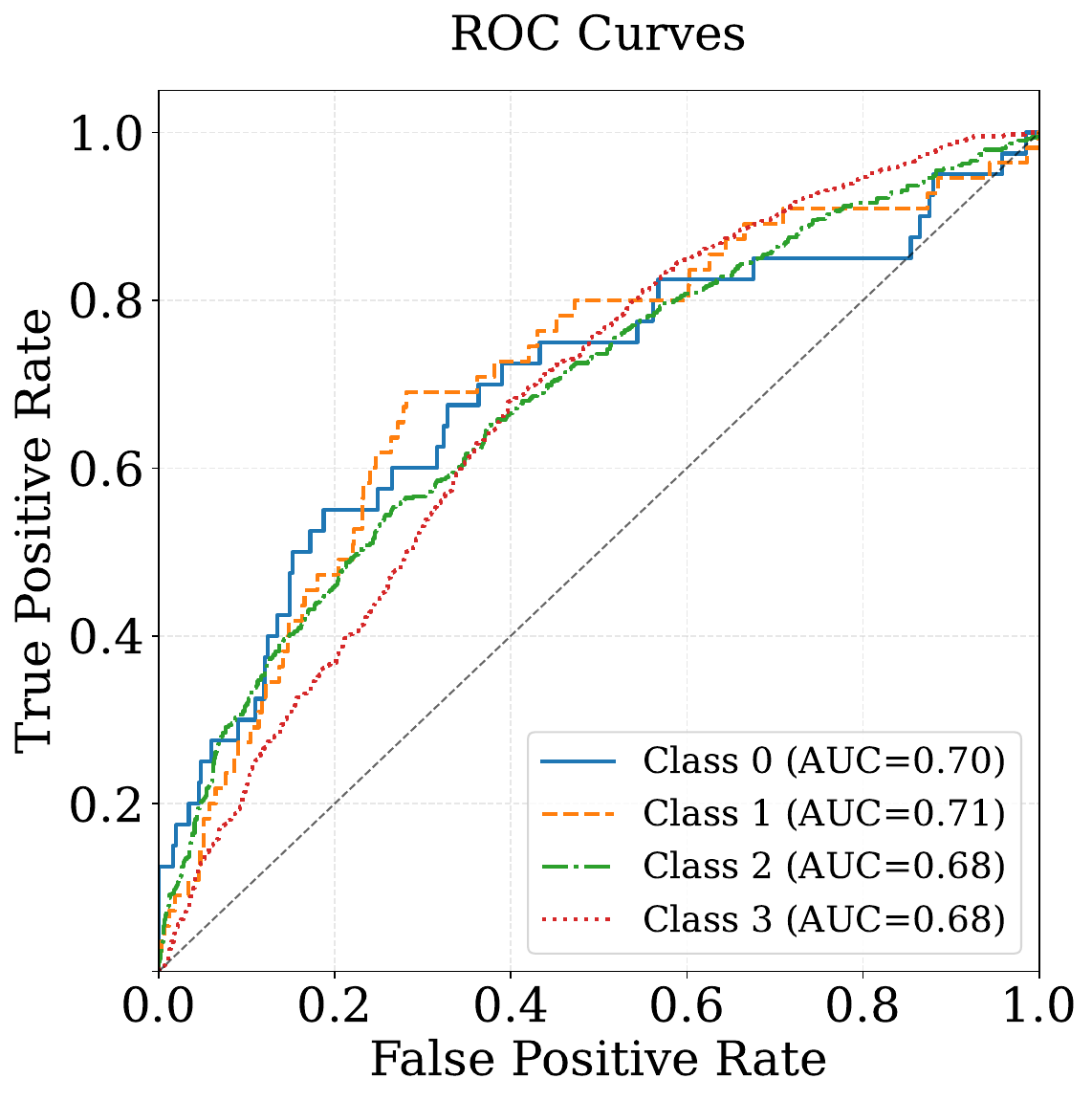}
    \caption{}
    \label{fig:roc_curve_hubert}
  \end{subfigure}
  \hfill
  \begin{subfigure}[b]{0.48\textwidth}
    \includegraphics[width=0.9\textwidth]{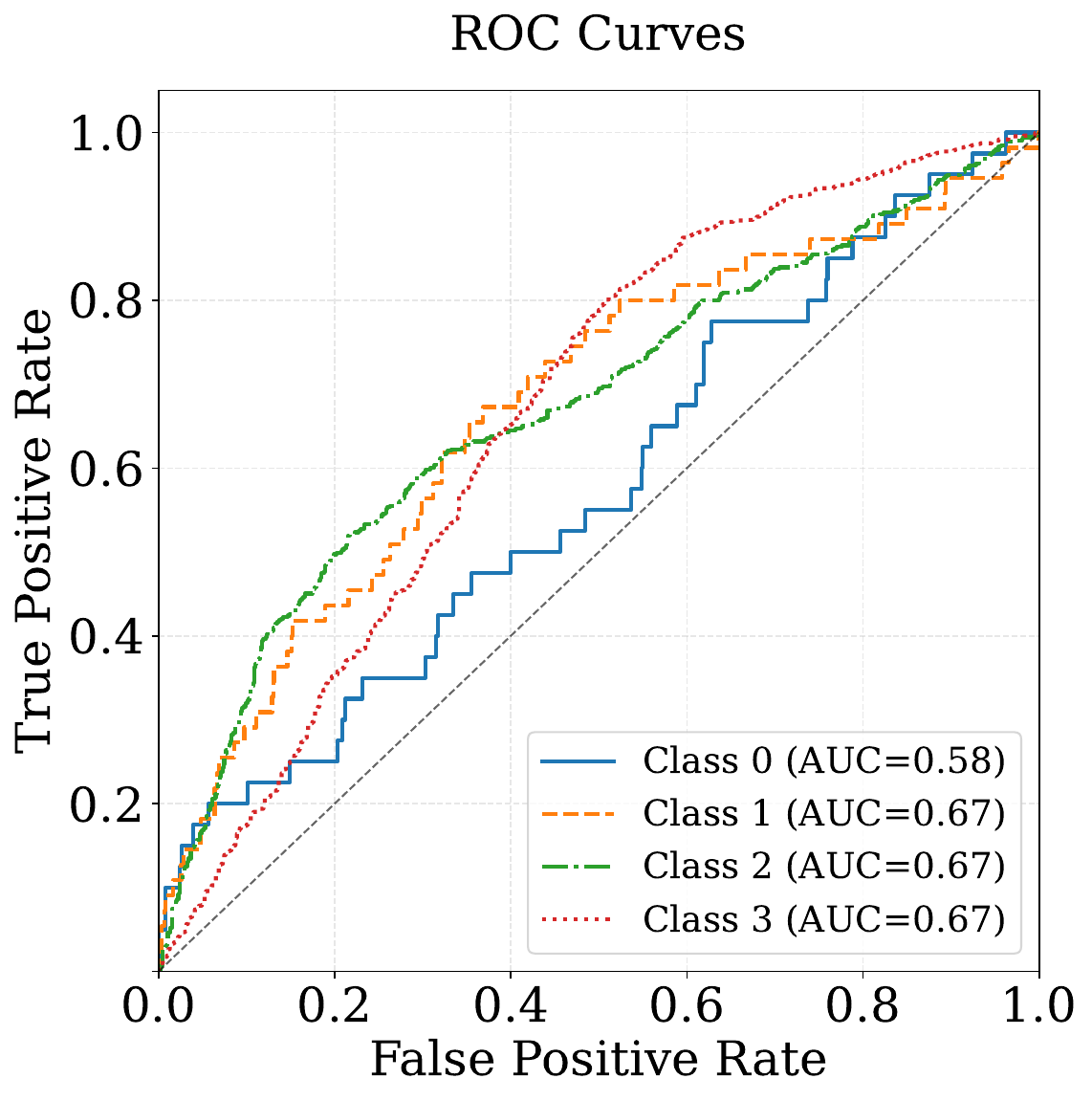}
    \caption{}
    \label{fig:roc_curve_bert_large}
  \end{subfigure}
  \begin{subfigure}[b]{0.48\textwidth}
    \includegraphics[width=0.9\textwidth]{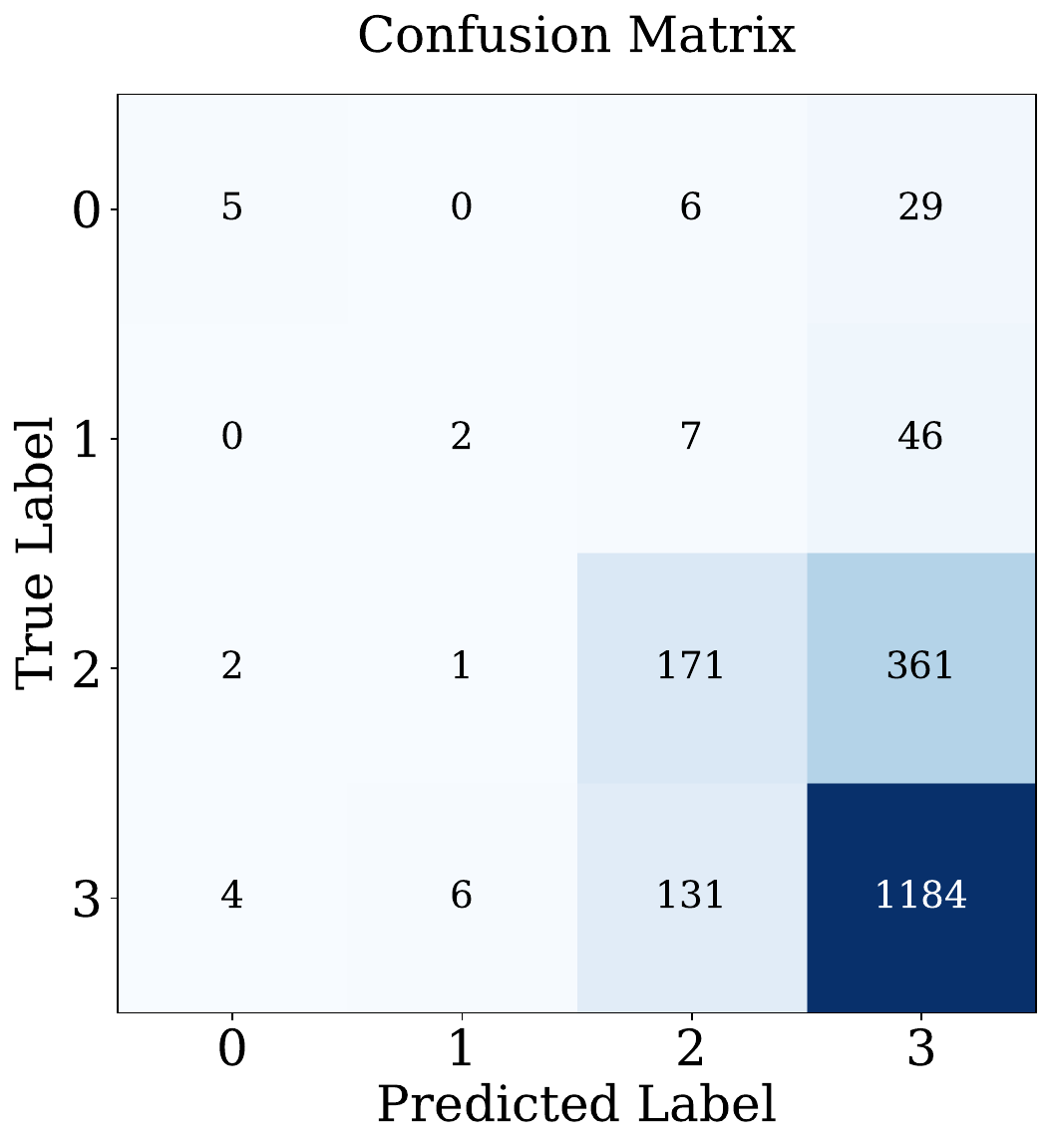}
    \caption{}
    \label{fig:confusion_matrix_hubert}
  \end{subfigure}
  \hfill
  \begin{subfigure}[b]{0.48\textwidth}
    \includegraphics[width=0.9\textwidth]{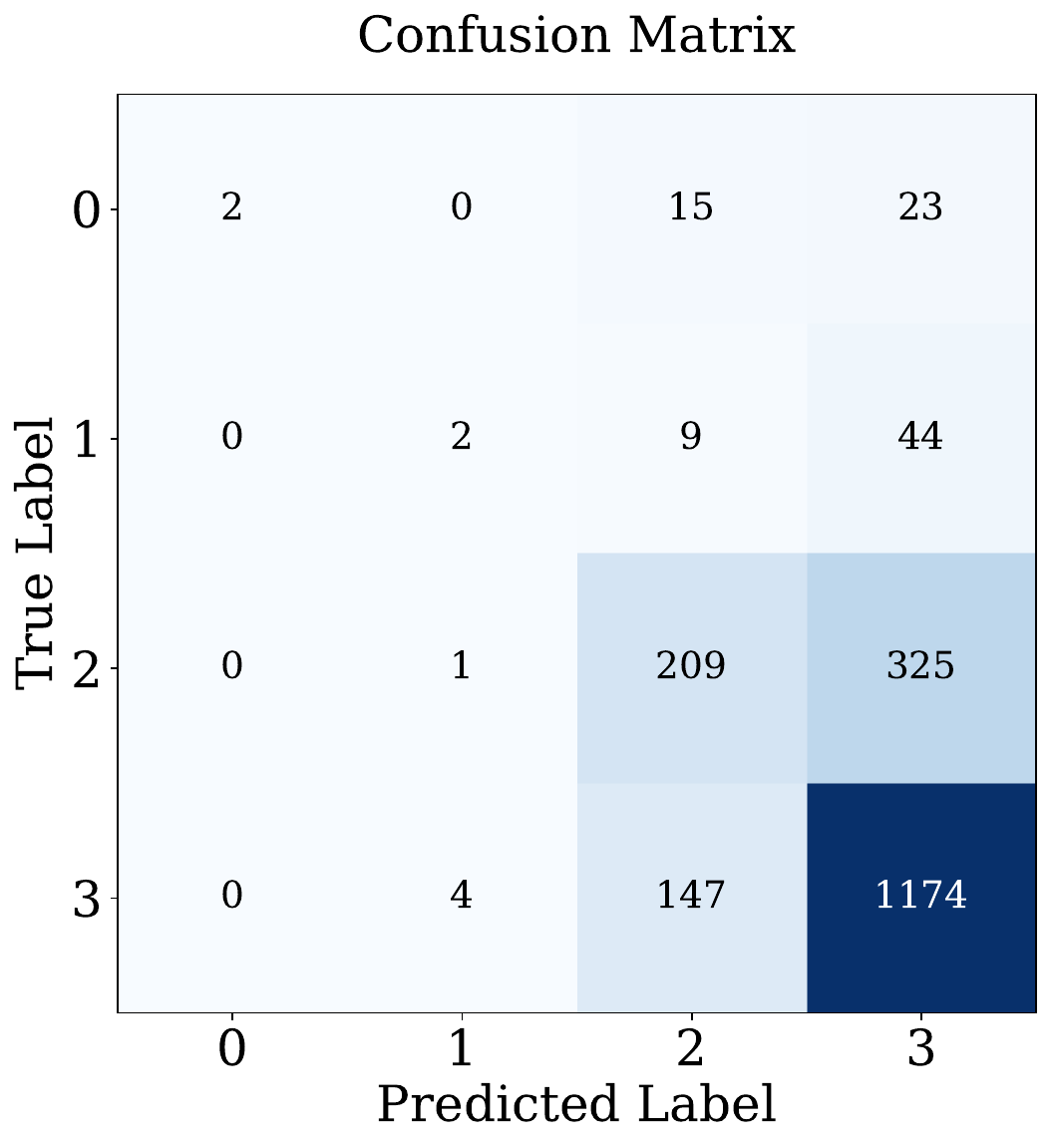}
    \caption{}
    \label{fig:confusion_matrix_bert_large}
  \end{subfigure}
  \caption{(a) ROC Curve of the Hubert model with backtranslation and focal loss and (b) 
   ROC Curve of the PULI-BERT-Large model with backtranslation and cross-entropy loss. (c) 
   Confusion Matrix of the Hubert model with backtranslation and focal loss and (d) 
   Confusion Matrix of the PULI-BERT-Large model with oversampling and cross-entropy loss. The results are averaged over 5 folds.}
  \label{fig:combined_transformer}
\end{figure*}

Despite PULI-BERT-Large being better in overall score, hubert shows higher minority-class sensitivity in the 
reported confusion matrices and ROC curves. A clear difference between
the two models can be observed in the ROC curves and confusion matrices in Fig.~\ref{fig:combined_transformer}.
The ROC curves show worse performance from both transformer models compared to the 
best Shallow Machine Learning models, however the transformer models are able to classify the minority classes better, 
as shown in the confusion matrices in Fig.~\ref{fig:confusion_matrix_hubert} and Fig.~\ref{fig:confusion_matrix_bert_large}.

\begin{table*}[!t]
    \centering
    
    \resizebox{1.0\textwidth}{!}{%
    \begin{tabular}{|c|c|c|c|c|}
    \hline
        \textbf{Losses} & \textbf{Accuracy} & \textbf{F1} & \textbf{ROC-AUC} & \textbf{Overall} \\ \hline
        \textbf{Cross-Entropy Loss} & \num{0.5614} & \num{0.5361} & \textbf{\num{0.6217}} & \textbf{\num{0.5731}} \\ \hline
        \textbf{Dice Loss} & \textbf{\num{0.5670}} & \textbf{\num{0.5529}} & \num{0.5951} & \num{0.5716} \\ \hline
        \textbf{Focal Loss} & \num{0.4800} & \num{0.4523} & \num{0.6029}& \num{0.5118} \\ \hline
        \textbf{Weighted Cross-Entropy Loss} & \num{0.4822} & \num{0.4502} & \num{0.6044} & \num{0.5123} \\ \hline
    \end{tabular}
    }
    \vspace{0.2cm}
    \caption{Results with different loss functions, metrics were averaged over 2 transformer models and different hyperparameters (like oversampling and data augmentation). The best results in each column are in bold.}
    \label{tab:transformer_loss_functions}
\end{table*}

The experiments with different loss functions are presented in Table~\ref{tab:transformer_loss_functions}. 
The best results were achieved by the Cross-Entropy Loss, which is the standard loss function for 
classification tasks, yields the highest aggregate performance among the tested losses. The Dice Loss achieved similar results, while the 
Focal loss and Weighted Cross-Entropy Loss achieved worse results.

\begin{table*}[!t]
    \centering
    \resizebox{1.0\textwidth}{!}{%
    \begin{tabular}{|c|c|c|c|c|}
    \hline
    \textbf{Method} & \textbf{Accuracy} & \textbf{F1} & \textbf{ROC-AUC} & \textbf{Overall} \\ \hline
    ROS & \num{0.5784} (\textcolor{green}{27\% $\uparrow$}) & \num{0.5914} (\textcolor{green}{30\% $\uparrow$}) & \num{0.6389} (\textcolor{green}{8\% $\uparrow$}) & \num{0.6029} (\textcolor{green}{22\% $\uparrow$}) \\ \hline
    Backtranslation & \num{0.6859} (\textcolor{green}{38\% $\uparrow$}) & \num{0.6119} (\textcolor{green}{32\% $\uparrow$}) & \num{0.6259} (\textcolor{green}{7\% $\uparrow$}) & \num{0.6412} (\textcolor{green}{26\% $\uparrow$}) \\ \hline
    Freeze Encoder & \num{0.4476} (\textcolor{red}{-15\% $\downarrow$}) & \num{0.4247} (\textcolor{red}{-14\% $\downarrow$}) & \num{0.5634} (\textcolor{red}{-8\% $\downarrow$}) & \num{0.4786} (\textcolor{red}{-12\% $\downarrow$}) \\ \hline
    \end{tabular}
    }
    \vspace{0.2cm}
    \caption{Ablation study results of different techniques on the transformer models using the 
    Hungarian essays dataset. The results are averaged over the 2 transformer models and different 
    hyperparameters. The values in parentheses indicate the relative change compared to the baseline 
    model without using the specific technique. Green values with upward arrows indicate improvement, 
    while red values with downward arrows indicate performance decrease.}
    \label{tab:transformer_ablations}
\end{table*}

The results of an ablation study on the transformer 
models are presented in Table~\ref{tab:transformer_ablations}.
The table shows the average performance of the models when 
using or not using a specific technique like oversampling, 
data augmentation and freezing the backbone during training.
As the table shows, both oversampling and data augmentation are 
associated with higher average transformer performance in this study.
Compared to shallow models, oversampling and data augmentation
have a more significant positive impact on the performance of the transformer models.
However, freezing the backbone during training significantly decreased
the performance of the models, which shows the importance of full fine-tuning
when adapting transformer models to specific tasks.

The overall results of the transformer models are
little lower than the best Shallow Machine Learning models. The best results with the transformer 
models are around $\approx$68\% overall score averaged over accuracy, F1-score, and ROC AUC metrics. 
However the transformer models are able to classify 
the minority classes better, as the confusion matrix in 
Fig.~\ref{fig:confusion_matrix_hubert} and Fig.~\ref{fig:confusion_matrix_bert_large} presents. 
Confusion matrices indicate improved minority-class detection for transformer models 
compared to shallow baselines in these settings. However,
due to the high class imbalance, most minority class samples are still misclassified 
as the majority classes.

\subsection{Statistical significance test}

In this section the results of the statistical significance test 
are presented. The T-test was used to compare the performance of 
the best shallow machine learning model (SMOTEBoost + Qwen3 
embeddings + Decision tree with 12 maximum depth) and the best 
transformer model (PULI-BERT-Large with backtranslation and 
cross-entropy loss) on the Hungarian essays dataset. The T-test was performed on the Accuracy, 
F1-score, ROC-AUC score and Macro-F1 score of the two models across the 5 folds.

The results show that there are no statistically significant differences 
between the two models in terms of Accuracy (t = \textbf{\num{-0.0464}}, 
p = \textbf{\num{0.9652}}) and Weighted F1-score (t = \textbf{\num{0.2862}}, 
p = \textbf{\num{0.7889}}).

Similarly, no statistically significant difference was observed for the 
Macro-F1 score (t = \textbf{\num{-0.9815}}, p = \textbf{\num{0.3819}}).

In contrast, a statistically significant difference was found for ROC-AUC 
(t = \textbf{\num{5.5053}}, p = \textbf{\num{0.0053}}), where the shallow 
model achieved higher scores than the transformer model.

\section{Discussion}
\label{sec:discussion}

The findings indicate that automatic reflective text classification 
in Hungarian achieves performance comparable to previously reported 
results in other languages \citep{zhang2024classification, nehyba2023applications}.
Shallow models obtained higher aggregate means, while statistical tests 
indicated no significant differences for Accuracy and Weighted F1.
The difference remains modest, suggesting that both 
paradigms are capable of capturing key characteristics of reflective
writing. While shallow models can distinguish between classes better,
transformer models show improved sensitivity to minority classes,
which is relevant for imbalanced educational datasets.
This study addresses document-level reflective text classification for Hungarian essays, 
highlighting the feasibility of this task in a morphologically 
rich language.

The findings further emphasize the importance of handling class 
imbalance. Techniques such as oversampling, class weighting, 
different loss functions, and data augmentation were particularly 
beneficial, especially for transformer-based models.
This suggests that transformer models are more sensitive 
to class distribution and can benefit substantially from 
strategies that improve the representation of minority classes.

These results may inform educational tools for preliminary feedback and workload 
support, with final assessment retained by educators. In particular, the ability of 
models to accurately 
assess reflective quality in writing can be leveraged in educational technologies, where automated 
feedback on student essays can enhance learning outcomes and the development of critical thinking skills.

Furthermore, these systems can reduce teachers' workload by providing initial assessments of 
student essays, reducing subjectivity and allowing educators to focus on more personalized 
feedback and instruction. The findings also suggest that integrating both shallow and deep learning 
approaches could yield hybrid models that use the strengths of each method, potentially leading to 
even more effective reflective analysis systems.

While our approach reports results for Hungarian reflective text classification, 
several limitations must be acknowledged. First, the effectiveness of these methods depend on class 
distribution and annotation quality. Even with oversampling, most models struggled to accurately classify minority 
classes, often achieving high overall accuracy by favoring majority class predictions. This highlights 
the need for further research into techniques that can better handle 
imbalanced datasets, particularly in the context of reflective analysis in Hungarian language essays. 
Future work could explore more advanced data augmentation strategies or the 
collection of more balanced datasets.

Another limitation arises from the restricted context window of transformer models, which makes 
the prediction more difficult due to the information loss when truncating longer essays into 
smaller chunks and aggregate them. Addressing this issue with Hungarian specific or multi 
lingual large language models that can handle longer contexts 
could further improve performance. 

\section{Conclusion}
\label{sec:conclusion}

This study examined automatic reflection-level classification in 
Hungarian student essays using a large expert-annotated corpus 
(N=1,954). This study provides a comprehensive document-level 
comparison in this Hungarian setting between shallow 
machine-learning methods and Hungarian pre-trained transformer 
models under severe class imbalance.

The main finding is a consistent trade-off between aggregate 
performance and minority-class behavior. Shallow models with 
strong feature representations achieved the best overall scores 
(approximately 71\% aggregate score over Accuracy, F1, and ROC-AUC), while 
transformer models achieved slightly lower overall performance 
(approximately 68\%) but provided more balanced behavior on 
underrepresented reflection levels. This indicates that, in 
low-resource and imbalanced educational NLP settings, 
classical approaches remain highly competitive, while 
transformer models are valuable when minority-class sensitivity 
is a priority.

Our experiments also show that imbalance-handling methods are 
not universally beneficial: their effect depends on model family 
and objective metric. In several configurations, these techniques 
improved minority-class detection but reduced majority-class 
stability, reinforcing the need for metric-aware model selection.

The study has two main limitations. First, the strong class skew 
remains a bottleneck for robust minority-class learning. Second, 
transformer context limits required chunking long essays, which 
can weaken document-level coherence signals. Future work should 
therefore focus on better minority-class data support (targeted 
collection and augmentation), longer-context Hungarian or 
multilingual architectures, and hybrid systems that combine 
the strong global discrimination of shallow models with the 
minority-class robustness of transformer-based methods.

\section*{List of abbreviations}

LLM - Large Language Model

\noindent OECD - Organisation for Economic Co-operation and Development

\noindent RWF - Reflective Writing Framework

\noindent PoS - Part of Speech

\noindent STEM - Science, Technology, Engineering and Mathematics

\noindent CS - Computer Science

\noindent NLP - Natural Language Processing

\noindent TF-IDF - Term Frequency-Inverse Document Frequency

\noindent RF - Random Forest

\noindent ROC-AUC - Receiver Operating Characteristic - Area Under Curve

\noindent Qwen3-E - Qwen3 Embeddings

\noindent OS - Oversampling

\noindent SMOTE - Synthetic Minority Over-sampling Technique

\noindent ROS - Random Oversampling

\section*{Declarations}

\subsection*{Data availability}
Due to ethical considerations, the datasets used and/or analyzed during the current study cannot be made publicly available and are not accessible.

\subsection*{Competing interests}
The authors declare that there are no conflicts of interest related to the research, authorship, and publication of this article.

\subsection*{Funding}

Multiple sources supported this research.
This research was supported by the Anonymous and the Anonymous grants.

% \subsection*{Authors' contributions

% \subsection*{Acknowledgements}

% Acknowledgments
% \section*{Acknowledgments}
% References - using Elsevier bibliography style
\bibliographystyle{cas-model2-names}
\bibliography{references}

\end{document}